\documentclass{article}

 \usepackage[final]{neurips_2025}
\usepackage{float} % add this in your preamble
\usepackage[utf8]{inputenc} % allow utf-8 input
\usepackage[T1]{fontenc}    % use 8-bit T1 fonts
\usepackage{hyperref}       % hyperlinks
\usepackage{url}            % simple URL typesetting
\usepackage{booktabs}       % professional-quality tables
\usepackage{amsfonts}       % blackboard math symbols
\usepackage{nicefrac}       % compact symbols for 1/2, etc.
\usepackage{microtype}      % microtypography
\usepackage{xcolor}         % colors
\usepackage{natbib}
\usepackage{hyperref}
\usepackage{pgfplots}
\usepackage{tikz}
\usepackage{subcaption}
\usepackage{graphicx}
\title{When Less is More: 8-bit Quantization Improves Continual Learning in Large Language Models}

\author{%
  Michael S. Zhang \\
  Algoverse\\
  \texttt{mzhang3518@gmail.com} \\
\And
  Rishi A. Ruia \\
  Algoverse \\
   \texttt{rishiru@outlook.com} \\
\And
  Arnav Kewalram \\
  Algoverse \\
   \texttt{arnav.kewalram@gmail.com} \\
\And
  Saathvik Dharmapuram \\
  Algoverse \\
   \texttt{saathvikd2686@gmail.com} \\
\And
  Utkarsh Sharma \\
  Algoverse \\
   \texttt{utkarsh@algoverseairesearch.org}
\And
  Kevin Zhu \\
  Algoverse \\
   \texttt{kevin@algoverseacademy.com}
}

\begin{document}

\maketitle

\begin{abstract}

Catastrophic forgetting poses a fundamental challenge in continual learning, particularly when models are quantized for deployment efficiency. We systematically investigate the interplay between quantization precision (FP16, INT8, INT4) and replay buffer strategies in large language models, revealing unexpected dynamics. While FP16 achieves superior initial task performance (74.44\% on NLU), we observe a striking inversion on subsequent tasks: quantized models outperform FP16 by 8-15\% on final task forward accuracy, with INT4 achieving nearly double FP16's performance on Code generation (40\% vs 20\%). Critically, even minimal replay buffers (0.1\%) dramatically improve retention—increasing NLU retention after Math training from 45\% to 65\% across all precision levels—with INT8 consistently achieving the optimal balance between learning plasticity and knowledge retention. We hypothesize that quantization-induced noise acts as implicit regularization, preventing the overfitting to new task gradients that plagues high-precision models. These findings challenge the conventional wisdom that higher precision is always preferable, suggesting instead that INT8 quantization offers both computational efficiency and superior continual learning dynamics. Our results provide practical guidelines for deploying compressed models in continual learning scenarios: small replay buffers (1-2\%) suffice for NLU tasks, while Math and Code benefit from moderate buffers (5-10\%, rising to 10-20\% for INT8/INT4 models), with quantized models requiring less replay than FP16 to achieve comparable retention. Code is available at https://github.com/Festyve/LessIsMore.
\end{abstract}

\section{Introduction}
Although large language models (LLMs) have achieved state-of-the-art performance across a range of natural language and reasoning tasks, their ability to retain knowledge over time remains a key limitation, particularly when models must be continually updated with new data. Scaling alone does not address this challenge, as repeated fine-tuning often causes older capabilities to deteriorate, a phenomenon known as catastrophic forgetting \cite{Luo_Yang_Meng_Li_Zhou_Zhang_2025, Carta_Cossu_Lomonaco_Bacciu_2021}. In real-world deployments, where models are expected to adapt continuously, this liability presents a major obstacle.
To solve this, researchers use replay-based methods.

The practical deployment of these models has necessitated considerable research focus on quantization, a compression methodology that transforms model parameters into low-precision representations \cite{Jacob2018, Dettmers_Lewis_Shleifer_Zettlemoyer_2022}. Quantization offers substantial gains in efficiency, enabling large models to train and run on commodity hardware, yet it also introduces new risks. Replay-based methods provide one of the most effective tools to mitigate forgetting. By selectively reintroducing samples from prior tasks during fine-tuning, replay buffers help stabilize model performance \cite{Chaudhry_Rohrbach_Elhoseiny_Ajanthan_Dokania_Torr_Ranzato_2019}. However, replay itself introduces another constraint: larger buffers improve retention but add computational and storage cost. This trade-off, however, remains underexplored, particularly for large language models.
This motivates our key research question:
\begin{enumerate}
    \item How does the trade-off between quantization precision and replay buffer size shape catastrophic forgetting in foundational models?
    % \item What is the Pareto Frontier between quantization precision and replay buffer size for retaining performance in continual learning?
\end{enumerate}
To answer this question, we fine-tune LLaMA-3.1-8B across three quantization levels (FP16, INT8, and INT4) using Low-Rank Adaptation \cite{Hu_Shen_Wallis_Allen-Zhu_Li_Wang_Wang_Chen_2021} for efficient adaptation, and evaluate on sequential tasks from the LoRI benchmark spanning natural language understanding, mathematical reasoning, and code generation \cite{Zhang_You_Panda_Goldstein_2025}. We vary replay buffer sizes (0\%, 0.1\%, 0.5\%, 1\%, 2\%, 5\%, 10\%, 20\%) of prior data to construct a quantization replay trade-off map, identifying where performance collapses and where minimal replay suffices to preserve accuracy.

Our results show that the cost of forgetting under quantization is not uniform: at higher precision, minimal replay is sufficient. Under INT4 quantization, buffer size becomes a decisive factor in preserving prior knowledge. Under INT8, quantization noise acts as a natural regularizer, enabling replay to be more effective even at modest buffer sizes, whereas under FP16 the absence of such noise makes the model more prone to overwriting prior knowledge despite replay, leading to steeper forgetting across tasks. From these findings, we provide empirical guidelines for balancing memory and accuracy in compressed continual learning and introduce a benchmark framework for future studies.
\section{Methods and Experimental Setup}
\label{sec:methods}
\paragraph{Model and quantization conditions.} We study a LLaMA-3.1-8B model in three precision levels: FP16, INT8, and INT4. All fine-tuning uses LoRA with rank $r=8$, alpha $a=16$ and 0.0 dropout across precisions; weights are frozen and adapters are learned. For INT8 we use standard inference-time weight quantization with LoRA adapters in higher precision; for INT4 we use QLoRA-style NF4 quantization \cite{Dettmers_Pagnoni_Holtzman_Zettlemoyer_2023}. The entire process is repeated independently for each precision. Following the LoRI benchmark protocol \cite{Zhang_You_Panda_Goldstein_2025}, we train for exactly 1 epoch per dataset to ensure comparability with prior work. All experiments use a learning rate of 2e-4, training batch size of 8, and evaluation batch size of 32. Runs were executed on a single NVIDIA B200 (180\,GB VRAM) with 28 vCPUs.
\paragraph{Tasks and datasets.} We partition continual tasks into (i) Natural Language Understanding (NLU): The model is trained on an aggregation of eight NLU datasets \cite{Hu_Wang_Lan_Xu_Lim_Bing_Xu_Poria_Lee_2023}, including BoolQ \cite{Clark_Lee_Chang_Kwiatkowski_Collins_Toutanova_2019}, PIQA \cite{Bisk_Zellers_LeBras_Gao_Choi_2020}, SocialIQA \cite{Sap_Rashkin_Chen_LeBras_Choi_2019}, ARC-Challenge \cite{Clark_Cowhey_Etzioni_Khot_Sabharwal_Schoenick_Tafjord_2018}, ARC-Easy \cite{Clark_Cowhey_Etzioni_Khot_Sabharwal_Schoenick_Tafjord_2018}, OpenBookQA \cite{Mihaylov_Clark_Khot_Sabharwal_2018}, HellaSwag \cite{Zellers_Holtzman_Bisk_Farhadi_Choi_2019}, and Winogrande \cite{Sakaguchi_Bras_Bhagavatula_Choi_2019}. We evaluate accuracy on the individual test split for each dataset. (ii) Mathematical Reasoning (Math): The model is trained on the GSM8K \cite{Cobbe_Kosaraju_Bavarian_Chen_Jun_Kaiser_Plappert_Tworek_Hilton_Nakano_et_al._2021} training split and evaluated on the GSM8K test split. (iii) Code Generation (Code): The model is trained on CodeAlpaca \cite{Chaudhary_2023} and evaluated using pass@1 on HumanEval \cite{Chen_Tworek_Jun_Yuan_Pinto_Kaplan_Edwards_Burda_Joseph_Brockman_et_al._2021}.

\paragraph{Continual learning with replay.}
Training proceeds in three stages, each followed by evaluation on all datasets (NLU, Math, Code):
\begin{itemize}\setlength\itemsep{0pt}
\item \textbf{Stage A} (initial): Train on 100\% NLU.

\item \textbf{Stage B} (second): Train on 100\% GSM8K while interleaving replay of size $B \in \{20,10,5,2,1,0.5,0.1,0\}\%$ for all NLU datasets respectively (uniform random sampling per dataset).

\item \textbf{Stage C} (third): Train on 100\% Code while interleaving replay of size $B \in \{20,10,5,2,1,0.5,0.1,0\}\%$ for all NLU and GSM8K datasets respectively (uniform random sampling per dataset).
\end{itemize}

\section{Analysis}

% Forward Accuracy

\begin{figure}[h]
\centering
    \includegraphics[width=15cm]{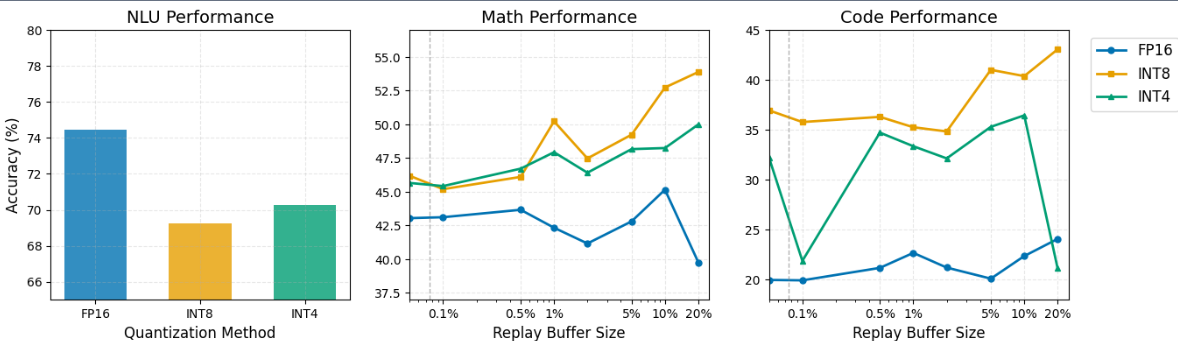}
    \caption{\textbf{Forward Accuracy.} Current task performance across different quantization levels and replay buffer sizes (log scale). NLU performance is unaffected by replay buffers.}
    \label{fig:forward_accuracy}
\end{figure}

\begin{figure}[htp]
    \centering
    \includegraphics[width=15cm]{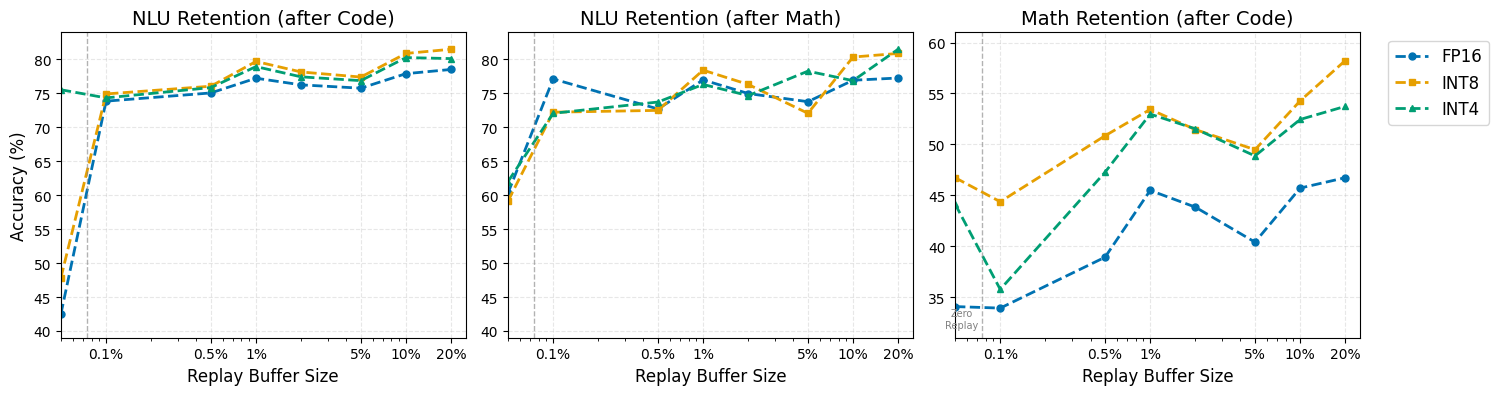}
    \caption{\textbf{Backward Accuracy.} Retention of previous task performance after training on subsequent tasks. Dashed lines indicate retention metrics. X-axis uses log scale to emphasize critical low-replay region.}
\label{fig:backward_accuracy}
\end{figure}

We analyze the effect of quantization and replay on continual learning performance, reporting both average forward accuracy (performance immediately after training on each task, Figure~\ref{fig:forward_accuracy}) and backward accuracy (retention of previous tasks, Figure~\ref{fig:backward_accuracy}).

\paragraph{Baseline degradation.} In isolation, quantization had only a minor effect on performance: INT8 models were within 1--2\% of FP16, and INT4 models degraded by 3--5\%. This confirms prior findings that quantization-aware training can preserve model accuracy at low bit-widths as demonstrated by \cite{Chen2023}.

\paragraph{Amplification under continual learning.} When tasks were learned sequentially without replay, these modest differences were amplified, with the high-precision FP16 model exhibiting the sharpest degradation. For example, under 0\% replay, FP16 NLU retention collapsed by roughly 30 points after subsequent Math and Code training---with several individual tasks falling toward chance level---whereas the quantized models, and INT4 in particular, retained substantially more of their prior NLU accuracy (Figure~\ref{fig:backward_accuracy}).

\paragraph{Replay as a stabilizer.}  Increasing the replay buffer size mitigated this amplification. At 20\% replay, the gap between INT4 and FP16 shrank back to single digits (INT4: 78.95\% vs FP16: 77.91\% average NLU). Thus, the critical finding is not that INT4 quantization is intrinsically harmful, but that its interaction with replay size determines long-term retention.

\paragraph{Interpretation.}
We posit that the unexpected performance gap between FP16 and the quantized models under replay may stem from how quantization noise interacts with continual learning dynamics. One possible explanation is that the high-precision FP16 model is more susceptible to overfitting on new task gradients. Its stability, while beneficial for the initial task, could allow new updates to overwrite prior knowledge more cleanly, thus exacerbating catastrophic forgetting.

In contrast, we hypothesize that INT8 and INT4 quantization may introduce a form of implicit regularization. This induced noise could smooth the loss landscape, potentially biasing the model towards flatter minima that generalize better across tasks. We speculate that this effect amplifies the relative influence of the few replayed samples, helping to anchor the model to previously learned knowledge and improving its ability to balance plasticity with stability. While the FP16 model begins with a higher initial accuracy, this proposed regularization effect could explain why its quantized counterparts ultimately demonstrate more robust knowledge retention and integration across the learning sequence. Further investigation is needed to validate this hypothesis.

\section{Related Works}
\label{sec:related_works}
\paragraph{Continual learning.} Continual learning aims to enable models to acquire new knowledge without erasing previous capabilities. However, due to parameter drift \cite{Luo_Yang_Meng_Li_Zhou_Zhang_2025}, models perform previously learned tasks inaccurately after being fine-tuned for new tasks. A large amount of the literature is focused on regularization-based methods \cite{Kirkpatrick_Pascanu_Rabinowitz_Veness_Desjardins_Rusu_Milan_Quan_Ramalho_Grabska-Barwinska_et_al_2017, Zenke_Poole_Ganguli_2017, Li_Hoiem_2017, Mallya_Davis_Lazebnik_2018}. These methods aim to constrain updates to important parameters during training on new tasks.  Although proven to be effective, these methods can often misidentify which weights are crucial, leading to unnecessary restriction or forgetting.

\paragraph{Replay sampling.} Replay-based methods inject stored examples from prior tasks during training to mitigate forgetting. Early strategies such as iCaRL \cite{Rebuffi_Kolesnikov_Sperl_Lampert_2017} used exemplar selection via herding, while Gradient Episodic Memory \cite{Lopez-Paz_Ranzato_2022} and A-GEM \cite{Chaudhry_Ranzato_Rohrbach_Elhoseiny_2019} enforced gradient constraints using stored data. Despite their effectiveness, nearly all replay methods assume full-precision training, leaving open the question of how buffer size and sampling strategies interact with quantized models. More recently, LifeQuant \cite{Chen2023} introduced lifelong quantization-aware training to stabilize knowledge retention during continual learning. LifeQuant largely focuses on highly quantized vision models, whereas our work focuses on large language models.

\paragraph{Quantization strategies.} To isolate the effects of reduced numerical precision, we employ various quantization strategies.
BitsAndBytes provides a practical and widely adopted library for applying 8-bit \cite{Dettmers_Lewis_Belkada_Zettlemoyer_2022} and 4-bit quantization to transformer-based models. It supports Quantization-Aware Training (QAT) and Post-Training Quantization (PTQ), enabling efficient model compression with minimal accuracy loss. \cite{Jacob2018} introduced QAT for convolutional models, while \cite{Banner2019} showed that 4-bit PTQ could be used effectively. \cite{Lin_Tang_Tang_Yang_Dang_Han_2023} extended low-bit quantization to LLMs through AWQ, adjusting for activation outliers to preserve accuracy at very low bit-widths. However, these studies were conducted in static training setups and do not address how quantization affects forgetting in models trained across multiple tasks.
Our work builds on these tools, specifically using BitsAndBytes to apply uniform quantization across 16, 8, and 4 bits, evaluating how different levels of compression impact catastrophic forgetting in continual learning.

\section{Conclusion and Limitations}
\label{sec:limitations}
We presented the first systematic study of how quantization and replay buffers interact during continual learning in large language models. Our experiments reveal that while higher-precision models require minimal replay to retain knowledge, aggressively quantized models are highly sensitive to buffer size. These findings suggest practical guidelines for deploying compressed models in real-world continual learning scenarios.

\paragraph{Recommendations.}
Based on the results, we recommend adopting a small replay buffer for NLU (1--2\%), which is sufficient across precisions; when stronger retention is required under INT8 or INT4 quantization, a moderate buffer (5--10\%) yields additional gains. For Math, allocate at least 5--10\% replay, increasing to 10--20\% for INT8/INT4 models when mathematical reasoning is a priority. For Code, INT4 models benefit from 5--10\% replay; enlarging the buffer to 10--20\% further improves stability in INT8 and FP16 settings.

\paragraph{Limitations.}
Our study has three main limitations. First, we restrict evaluation to LLaMA-3.1-8B and a limited set of tasks (NLU, Math, Code), which may not generalize to other architectures or domains such as multi-modal learning. Second, we consider only a limited set of uniform quantization levels (FP16, INT8, INT4) and do not explore mixed-precision or adaptive quantization strategies. Furthermore, our experiments lack multi-seed runs and confidence intervals. Finally, replay was implemented with simple reservoir sampling; more sophisticated selection strategies (e.g., herding, clustering) may further shift the trade-offs. Future work should extend our benchmark to broader models, adaptive quantization, and alternative replay mechanisms.

\begin{ack}
This paper was supported by Algoverse and would not have been possible without the mentorship of Utkarsh Sharma.
\end{ack}

\bibliographystyle{plain}
\bibliography{citations}

\end{document}